# An ensemble of machine learning and anti-learning methods for predicting tumour patient survival rates.

Christopher Roadknight
Horizon Digital Economy Research
School of Computer Science
University of Nottingham
Chris.roadknight@nottingham.ac.uk

Durga Suryanarayanan
University of Nottingham Malaysia Campus, School of Computer Science
University of Nottingham
Durga.Suryanarayanan@nottingham.edu,my

Uwe Aickelin
School of Computer Science
University of Nottingham
Uwe.Aickelin@nottingham.ac.uk

John Scholefield
Faculty of Medicine & Health Sciences
University of Nottingham
John.Scolefield@nottingham.ac.uk

Lindy Durrant
Faculty of Medicine & Health Sciences
University of Nottingham
Lindy.Durrant@nottingham.ac.uk

*Abstract*— **This paper primarily addresses a dataset relating to cellular, chemical and physical conditions of patients gathered at the time they are operated upon to remove colorectal tumours. This data provides a unique insight into the biochemical and immunological status of patients at the point of tumour removal along with information about tumour classification and post-operative survival. The relationship between severity of tumour, based on TNM staging, and survival is still unclear for patients with TNM stage 2 and 3 tumours. We ask whether it is possible to predict survival rate more accurately using a selection of machine learning techniques applied to subsets of data to gain a deeper understanding of the relationships between a patient's biochemical markers and survival.**

**We use a range of feature selection and single classification techniques to predict the 5 year survival rate of TNM stage 2 and 3 patients which initially produces less than ideal results. The performance of each model individually is then compared with subsets of the data where agreement is reached for multiple models. This novel method of selective ensembling demonstrates that significant improvements in model accuracy on an unseen test set can be achieved for patients where agreement between models is achieved. Finally we point at a possible method to identify whether a patients prognosis can be accurately predicted or not.**

*Keywords—Ensemble, Bioinformatics, Machine Learning.*

1. INTRODUCTION

Colorectal cancer, commonly known as colon cancer or bowel cancer, is a form of cancer due to uncontrolled cell growth in the colon or rectum (parts of the large intestine), or in the appendix [1]; and is third most common cancer in men (663 000 cases, 10.0% of the total) and the second in women (571 000 cases, 9.4% of the total) worldwide. Around 60% of cases were diagnosed in the developed world. It is estimated that worldwide, in 2008, 1.23 million new cases of colorectal cancer were clinically diagnosed, and that there were 608,000 deaths due to this form of cancer, making it the fourth most common cause of death due to cancer [2].

The colon or rectal cancers begin in the digestive system (also known as the gastro-intestinal system). The wall of colon and rectum are made up of layers of tissues. The cancer starts in the inner layer and grows through the other layers underneath through the bowel wall. The extent of a person's cancer is commonly described using a TNM Cancer Staging system. The TNM staging system for all solid tumours was devised by Pierre Denoix to classify the progression of cancer, through the following three attributes:

1. T describes the size of primary (original) Tumor and whether it has invaded nearby tissues
2. N describes the involvement of nearby lymph Nodes
3. M describes the spreading of cancer to other distant parts of the body (Metastasis)

Based on the TNM stage and other biological factors such as age and health, subsequent treatment options are planned. However, patient prognosis is currently a poorly understood process. Though patients with a TNM stage of 1 or 4 have strong distinction with regard to predicting the survival period, those with TNM 2 & 3 stages have very poor prognosis accuracy, thus making TNM Stage a poor marker for prognosis. One current approach for predicting better survival rates combines the results of applying learning and anti-learning algorithms on a variety of physical, immunological, biochemical and clinical data collected from the patients after the tumour had been removed, combined with the TNM stage, and has been proven to yield better prognosis [3]. Anti-learning is the term given for the situation where performance of a trained computational intelligence technique is significantly worse than random guessing and is not overfitting or overtraining [4]. Anti-learning has been observed in a range of synthetic and real-world datasets [eg. 16, 17, 18, 27]

The data for this research was gathered by scientists and clinicians at City Hospital, Nottingham. The dataset we use here is made up of over 200 possible attributes for 462 patients. The attributes are generated by recording metrics at the time of tumour removal, these include:

- Physical data (age, sex etc)
- Immunological data (levels of various T cell subsets)
- Biochemical data (levels of certain proteins)
- Retrospective data (post-operative survival statistics)
- Clinical data (Tumour location, size etc.).

In the research into the relationship between immune response and tumour staging there has been some support of the hypothesis that the adaptive immune response influences the behaviour of human tumours. In situ analysis of tumour-infiltrating immune cells may therefore be a valuable prognostic tool in the treatment of colorectal cancer [5]. The immune and inflammation responses appear to have a role to play in the responses of patients to cancer [6] but the precise nature of this is still unclear.

This research initially attempts to characterize the dataset and inter-relationships between attributes and then models cause effect relationships within the data using single and ensemble methods. We show that these tasks are extremely difficult using conventional techniques and that the dataset might belong to a subset of dataset that require a unique approach. We do this by using a range of supervised methods in an attempt to characterize features of the dataset and accommodate inadequacies such as missing data.

## 2. MACHINE LEARNING AND ANTI-LEARNING TO predict TNM Stage FROM PATIENT DATA

The meaningfulness of nearest neighbour style clustering in highly dimensional data has be discussed previously [7] and it can be argued that there are serious problems with using this approach alone. [8]. A supervised approach has more scope for reducing the dimensionality within the algorithm. It is relatively trivial to build a model that best fits the data, even with numerous attributes and missing values.

Unfortunately this model is very likely to be memorising unique combinations of values for each patient. This is why models are tested on an unseen test set to decide how well the trained model generalises to the "rest of the world". Here we attempt to build a testable model that predicts the TNM stage of a patient from the other available data. If we can more clearly define the relationship between physical TNM staging and the biochemical state of a patient we can hypothesise that treating a patient's biochemical state with methods such as chemotherapy, immunotherapy and radiotherapy based on TNM stage is valid.

Several methods were used in an attempt to predict the appropriate TNM stage of a patient from their attribute set. The methods used included Bayesian Networks [9], Naïve Bayes Classifier [10], Classification and Regression Trees (CART) [11], Multilayer Perceptron [12] and Support Vector Machines (SVM) [13]. These were either self-programmed, available in the R packages or the WEKA toolkit [14] or used other existing software suites [15].

When initially looking at all 4 TNM stages there was some success at predicting stages from the attribute set, particularly when some of the patients and attributes with the most missing data were removed. Most success was achieved when predicting TNM stage 1 and 4, which were the least and most severe stages respectively [16]. Differentiating between TNM stage 2 and 3 using available patient data was a much more complicated task but modelling the relationship between patients data and TNM these 2 important tumour stages was best achieved using the inverse of the model prediction and exploiting the anti-learning phenomena. We compared a range of classifiers and attribute selection methods and consistently found test set performances of worse than guessing. It must be remembered that approaches were optimized for test set performance and performance of the training set was much better (80-98% accurate). This kind of behaviour is rare but dataset types it has been observed in include biological data in general and cancer data in particular [28].

We arrive at the following set of facts:
➢ Some functions are best modelled using a machine learning approach
➢ Some functions are best modelling using a machine anti-learning approach
➢ Some cause-effect relationships are the aggregate of many individual functions

It is logical to therefore conclude that some data modelling tasks will be ideally implemented using both learning and anti-learning approaches on relevant subsets of the data. For instance, both hadamard matrices [17] and simple polynomial functions (eg. $A=x+(1/y)-z$) are 100% solvable by simple anti-learning and learning methods (eg. SVM, Multilayer Perceptron) respectively but if you merge these datasets and attempt to predict the XOR of the 2 outputs results fall dramatically to approximately 50% regardless of if you take an anti-learning or learning approach.

### 3. SUPERVISED ENSEMBLE LEARNING OF SURVIVAL RATES

We have detailed information about post-operative survival, so we can also model how attribute values effect survival rates. Several of the physical attributes available in the dataset pertain to the survival of the patients after their operation to remove the tumour. The number of months the patient has survived, whether they are still alive or not and how they died are all available. Figure 1 shows survival curves for patients up to 60 months, after which, clinicians deem they have 'survived' their colorectal tumour. The strong difference between survival rates in TNM stage 1 and 4 patients is apparent (i.e. at 30 month the survival rate is approximately 95% and 5% for these 2 groups). The difference between patients with TNM stage 2 and 3 cancers is less apparent.

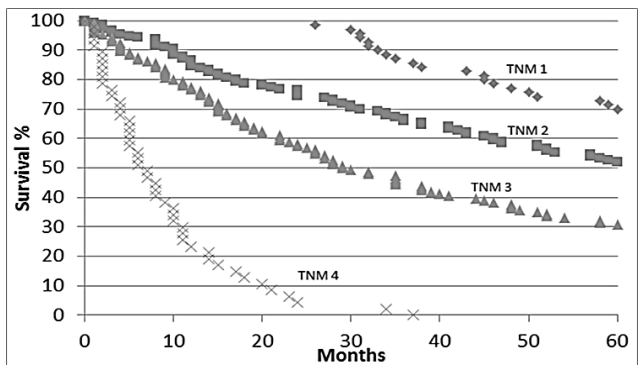

Figure 1. Survival Curves for patients at all 4 TNM stages

Again focusing on just TNM stage 2 and 3 patients we attempted to predict survival using both AI techniques and the TNM stage itself. The term "survival" is somewhat subjective but for the purposes of this work we used 5 years as the threshold for survival. Using the TNM stage alone to predict survival gives an accuracy of 64.6% (155 correct from 240). This is achieved by stating that all type 2 tumour patients will survive and all type 3 will not.

Due to the high dimensionality of the data using the entire dataset yields very poor results [16] so a degree of attribute selection is required. To achieve this, the dataset was ranked based on four evaluators, and the top ten attributes were used in training various models. The results of the experiments are shown in Figure 2. Accuracy is used here and throughout this paper as it is the most transparent measurement when positive and negative outcomes are close to equivalent, as they are here. Also, all results shown are for the unseen test data via 10 fold cross validation. It can be clearly seen that the SVM ranking [19] works best across various algorithms. For SMO and PegaSOS (both based on Support Vector Machines), SVM Ranking is clearly the better ranker. However, it is interesting to note that the SVM Ranked attributes perform best in decision trees, and other algorithms as well. The ChiSquared [20] and InfoGain [21] attribute evaluators have the same best 10 attributes, but ranked in a different order. However, for the learner, these two rankings are the same and hence they have the same prediction accuracies. Based on the graph in Figure 2 and the trial experiments, SVM ranking emerged undisputedly as the best ranking algorithm for this sample dataset. The best 10 attributes and the worst 10 attributes based on SVM Ranking are listed in Table 1, it was felt that taking the highest and lowest 10 values was sufficient to create subsets. A detailed description what these markers represent is beyond the scope of this paper but all have some relevance to a patients cancer situation.

Experiments were conducted with the SVM Ranked data sample, to decide on the algorithms to work as an ensemble. The final list of algorithms used is –

- SVM (Sequential Minimal Optimization) [22]
- Logistic Regression [23]
- Classification and Regression Trees [11]
- J48 trees [24]

Experiments were also conducted to learn on the anti-learnable properties of the data.

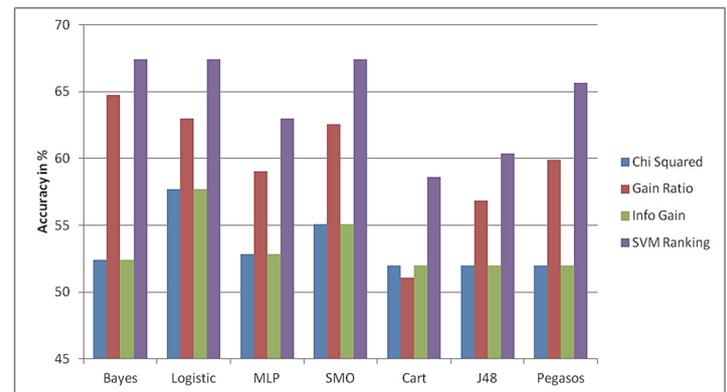

Figure 2. Comparison of Evaluators

Using the SVM (SMO) algorithm as an example, the best 10 and the worst 10 attributes were used to train the machine learning algorithm, using a leave-one-out cross validation. The results for various algorithms on using the worst 10 (table 2) and best 10 (table 3) attributes are tabulated below. The primary objective of this exercise is to find the optimum number of attributes that can be used for training the learning algorithms.

TABLE 1
BEST AND WORST TEN ATTRIBUTES BASED ON SVM RANKING

| Best 10 attributes | Worst 10 attributes |
|---|---|
| age | pERK |
| IL17tc | FLIP |
| p27cyto | ki67 |
| nucint2 | cxcl9lowhigh |
| Tstroma | cd46 |
| CXCR4xtilelovsho | micarec |
| ulbp23 | socs1 |
| cCD24ordrec | betcytop |
| IRF2 | betmembr |
| statcyto.int | FLIPsCyto |

TABLE 2
SVM (SMO) ALGORITHM WORST PERFORMING ATTRIBUTES

TABLE 3
SVM (SMO) ALGORITHM BEST PERFORMING ATTRIBUTES

| Number of Attributes | SVM Anti-Learning |
|---|---|
| 10 | 43.33 |
| 9 | 45.83 |
| 8 | 41.66 |
| 7 | 47.91 |
| 6 | 45 |
| 5 | 50.41 |
| 4 | 47.08 |
| 3 | 48.75 |
| 2 | 51.25 |
| 1 | 52.91 |

Using the best 8 attributes clearly gives maximum prediction accuracy on the dataset. Using the worst 8

| Number of Attributes | SVM Learning |
|---|---|
| 10 | 62.08 |
| 9 | 63.33 |
| 8 | 64.58 |
| 7 | 62.5 |
| 6 | 62.5 |
| 5 | 62.08 |
| 4 | 62.08 |
| 3 | 57.08 |
| 2 | 60 |
| 1 | 49 |

attributes, SVM gives the worst prediction accuracy of 41.66%, which, on performing the negation, will lead to the maximum prediction accuracy. Similar experiments were carried out using other algorithms to decide the optimal number of attributes. The final list of algorithms is –

- Simple CART – TNM Staging
- SVM(SMO) – Best 8 attributes
- SVM(SMO) – Worst 8 attributes
- Logistic – Best 8 attributes
- Logistic – Worst 9 attributes
- J48 – Best 8 attributes

The ensemble has been constructed to use a logical AND function to vote on the class of the input sample, this is mathematically trivial but allows for novel and powerful aggregation of decisions. The ensemble consists of all possible subsets for 6 algorithms. This gives a total of 63 possible combinations, inclusive of single algorithm results and a combination of all 6 algorithms. The 63 combinations consists of the following subsets –

- 6 – single algorithms
- 15 – 2 algorithm combinations
- 20 – 3 algorithm combinations
- 15 – 4 algorithm combinations
- 6 – 5 algorithm combinations
- 1 – combination of all 6 algorithms

Figure 3 displays the prediction accuracies of the ensemble alongside the number of matches for that particular ensemble, with the number of algorithms used on the X axis. In this case 'number of matches' means how many patients achieved agreement on the outcome. Obviously when only 1 algorithm is used all patients are included, as more algorithms are compared, agreement is less usual. It is to be expected that the number of matching results decreases in a regular fashion as the number of algorithms increase in the ensemble as complete agreement becomes less likely. It is also observable that as the number of algorithms in the ensemble increases there is a general increase in prediction accuracy from approximately 60-65% to nearly 90%. This increase in accuracy is accompanied by a decrease in the number of patients this accuracy is achievable for.

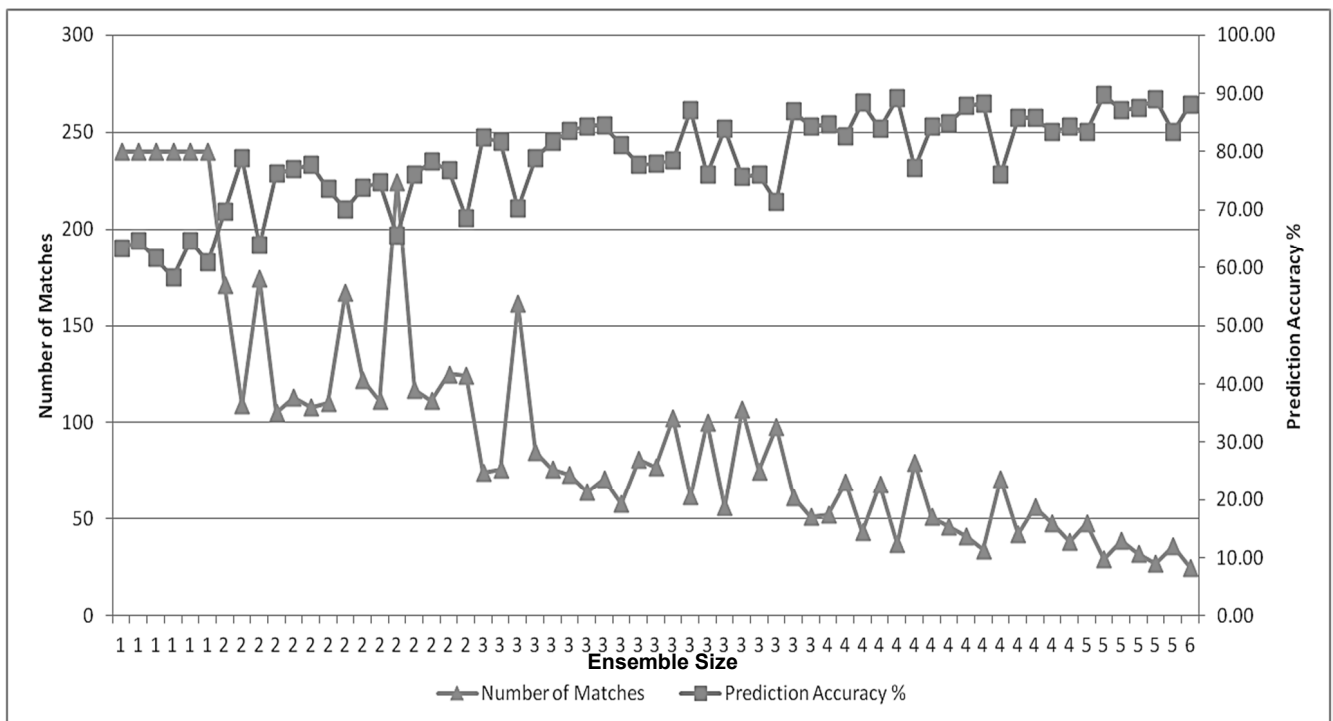

Figure 3. Number of Matches and Prediction Accuracy vs Ensemble size

A graph of the prediction accuracies of the various ensembles is presented in Figure 4. The highest prediction accuracy of 89.66% is achieved by combining all algorithms apart from the predictions of best 8 attributes using SVM. However, as shown in figure 3 the five above mentioned algorithms agree for only 29 data samples.

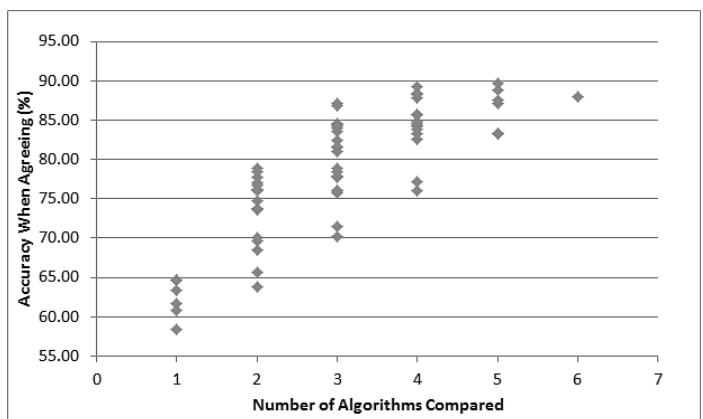

Figure 4. Prediction Accuracies of Ensemble

As a final experiment we looked at the biomarker statistics for patients who had complete agreement for all six algorithms and compared them to patients where the decision on prognosis was most diverse, namely, 3 algorithms predicted survival **over** 5 years and 3 predicted survival **under** 5 years. Those where there was complete agreement of algorithms were deemed to have a higher 'ease of prognosis' value. The best single biomarker for predicting the 'ease of prognosis' was ulbp3rec. ULBPs activate signaling pathways in primary Natural Killer cells, resulting in the production of cytokines and chemokines. It appears that when this variant is present a patient's survival rate is more difficult to predict. Based on a 0 or 1 scoring mechanism, ulbp3rec is present at an average rate of **0.6111** on patients that our ensemble couldn't agree on vs **0.1877** for patients where all six ensemble algorithms agree. Identifying 'difficult' patients using this method is about **72%** accurate. If we add two other markers (Tstroma and p53) and use a simple MLP we can achieve a slightly higher **83%** accuracy in predicting 'difficult' patients. This final analysis is on a relatively small number of patients (36) and is indicative rather than conclusive at this stage.

To show how new survival predictions models compare to conventional methods of survival prediction, the dataset can be divided into 4 distinct subgroups, each with a corresponding survival rate:

1. TNM 2 patients predicted to survive by the ensemble learning model. (TNM 2/model = survive)
2. TNM 2 patients predicted not to survive by the ensemble learning model (TNM 2/model = not survive)
3. TNM 3 patients predicted to survive by the ensemble learning model (TNM 3/model = survive)
4. TNM 3 patients predicted not to survive by the ensemble learning model (TNM 3/model = not survive)

One interesting observation about this is that survival of TNM stage 3 patients with a positive prognosis from the model have a better survival rate than TNM 2 patients with a negative prognosis. This is shown more clearly by plotting the 4 corresponding survival curves (fig. 5)

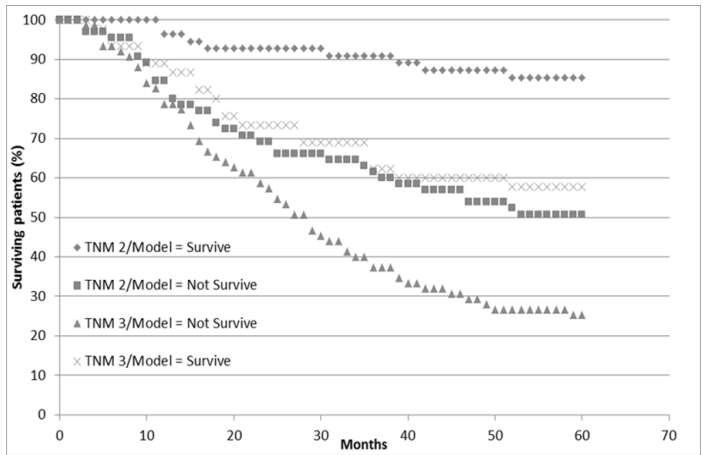

Figure 5. Survival curves for 4 distinct patient sets grouped on TNM stage and model prognosis

## 4. DISCUSSION

We have presented results of a systematic analysis of a unique dataset based on a range of factors associated with colorectal tumour patients. This dataset is limited in many ways, but extremely important nonetheless and modeling any relationships or features based on the dataset to hand is an urgent priority. Generally, whether attempting to predict TNM stages or survival, patients at TNM stage 1 and 4 have more clear indicators in the attribute set. TNM stage 2 and 3 patients provide a much more challenging prediction task, so much so that the TNM stage appears much less important when predicting survival for these 2 stages than other indicators. The fact that prediction of survival can be achieved at comparable rates to TNM staging suggests that survival is actually based on BOTH the physical metrics used for TNM staging AND the biochemical and immunological markers presented at the time of tumorectomy.

Rule tree, Bayesian and Neural approaches have been used with some limited success for prediction, but in most experiments there is a lack of repeatable success in developing a model that accurately predicts survival or TNM stage on an unseen test set. One possible reason for this could be over-fitting, though a well-constructed ANN or CART tree shouldn't exhibit over fitting and in any case they shouldn't be WORSE than guessing. Another possibility is poor or inaccurate labelling of patients tumour stages. But again this should only result in poor performance on the unseen test set. Methods such as Correlated Activity Pruning [25] may be useful in ensuring a minimal sized model and will be one focus of future research. There might be improvement to learning by using recent advancements such as multiple kernel learning [26] but it is just as likely, as with boosting, methods that improve learning may be just as effective at amplifying values in the opposite direction. Overall this is an iterative process with a large number of steps, each providing more insight into the dataset and its modeling. There is substantial pre-processing required and it is significant that pre-processing the patient's data (selecting based on different thresholds) has a significant difference on the resulting models.

This failure to accurately classify TNM stages or survival periods may in fact be useful if we suspect there are subsets within the groups. The failure to correctly classify a set of patients may mean these patients have different characteristics while still expressing the same classification of tumour. This would imply that treatment based solely on tumour classification would be sub-optimal.

We have proposed an explanation for the results which is a phenomenon called "Anti-learning". Here, unique characteristics of the dataset lead to a condition where validation on an unseen test set produces results significantly and repeatedly worse than guessing. Interestingly, one real world dataset that demonstrates this behavior is very similar to the dataset used here, being the classification of response to chemo/radiotherapy in Esophageal Adenocarcinoma patients using microarray data of biopsied patients [27] [28]. It is possible to then infer that with some highly dimensional complex biological data sets, when we have a relatively small sample size, anti-learning exists. Initial experiments appeared to show that the best possible approach to classifying patients with TNM stage 2 and 3 tumours was to focus on anti-optimizing the learning process to achieve the worse possible test set performance and then inverting the underlying model. Overall when looking for test set performance on the important TNM stage 2 and 3 patients, the best possible results can be achieved if we inverted the answer supplied by an ADABoosted SVM or ANN. Using these methods it is possible to achieve reliable prediction rates on an unseen test set of higher than any learning algorithm attempted. It is not impossible to imagine that many complex biological datasets also present us with a small, noisy sample of a much bigger complex dataset and this must be investigated further. The dataset used here has several drawbacks when it comes to building robust cause-effect models. The 2 main issues are the missing values and the high dimensionality.

By pre-processing the dataset to convert attributes that may have a non-linear effect into linear attributes, missing values can be represented as means of all values with a more stable meaning. This pre-processing step also makes resulting machine learning approaches more capable of reaching global minima. The second issue of high dimensionality is tackled by using an attribute selection approach first demonstrated on another cancer dataset. We not only use the most highly ranked attributes here but by using anti-learning on the lowest ranked attributes we can produce other, distinct models of the data.

When looking specifically at survival, using 3-6 unique predictive approaches allows us to compare predictions from all these approaches. The significant amount of differences between the predictions for these approaches could be inferred from the lack of relationships between TNM staging and the immunohistochemistry [16] and the lack of a relationship between unsupervised clusters and TNM stages

[7]. The introduction of anti-learning methods to the ensemble approach also makes for a diverse set of algorithms capable of modeling complex relationships. So for the set of patients, the level of agreement can fall to as low as 10% but when agreement occurred, predictive performance increased up to a maximum of 90%. This level of accuracy for predicting survival of patients with TNM stage 2 and 3 tumours is unprecedented, albeit on a small subset of patients.

## 5. CONCLUSIONS

In this paper we show that the relationship between clinical markers, physical tumour characteristics and patient survival is a complex one, but using a range of preprocessing, machine learning and ensemble learning approaches we make significant improvements in the predictability of survival. The ensemble learning approach offers an important opportunity for clinicians to improve and customise prognosis for patients but also, any information gained by analysing the agreeing models can be fed back to researchers in their endeavors to improve treatment options at a cellular level. It is apparent that small changes to the pre-processing mechanisms can yield significant changes to the ensemble learning accuracy but due to the highly dimensional nature of the data, optimising and automating the pre-processing and feature selection stage is currently a subject of further research. We also show for the first time that inverse ranking methods are an efficient method for attribute selection for anti-learning methods. Our ensemble learning models show the highest accuracy so far seen in predicting survival in patients, albeit a subset of patients.

Real world datasets sometimes consist of multiple relationships with differing degrees of learnability. If we can deconstruct these datasets into solvable sub-components we will be making a huge leap in tackling real world complex and big data modelling tasks.

## ACKNOWLEDGEMENTS

This work was in part supported by the Neo-demographics project funded by EPSRC (EP/L021080/1)